\documentclass{article}

\usepackage[preprint]{neurips_2023}

\usepackage[utf8]{inputenc} 
\usepackage[T1]{fontenc}    
\usepackage{hyperref}       
\usepackage{url}            
\usepackage{booktabs}       
\usepackage{amsfonts}       
\usepackage{nicefrac}       
\usepackage{microtype}      
\usepackage{xcolor}         

\usepackage{graphicx}
\usepackage{multirow}
\usepackage{csquotes}
\usepackage{subcaption}
\usepackage{lineno}

\usepackage{enumitem, hyperref}
\makeatletter
\def\namedlabel#1#2{\begingroup
    #2%
    \def\@currentlabel{#2}%
    \phantomsection\label{#1}\endgroup
}
\makeatother

\title{Evaluating the Effectiveness of Retrieval-Augmented Large Language Models in Scientific Document Reasoning}

\author{%
  Sai Munikoti\thanks{These authors contributed equally to this work.},~ Anurag Acharya, Sridevi Wagle, Sameera Horawalavithana\footnotemark[1]\\
  Pacific Northwest National Laboratory\\
  \texttt{\{sai.munikoti,anurag.acharya,sridevi.wagle,yasanka.horawalavithana\}@pnnl.gov} \\
}

\begin{document}

\maketitle

\begin{abstract}
Despite the dramatic progress in Large Language Model (LLM) development, LLMs often provide seemingly plausible but not factual information, often referred to as hallucinations. 
Retrieval-augmented LLMs provide a non-parametric approach to solve these issues by retrieving relevant information from external data sources and augment the training process. These models help to trace evidence from an externally provided knowledge base allowing the model predictions to be better interpreted and verified. In this work, we critically evaluate these models in their ability to perform in scientific document reasoning tasks. To this end, we tuned multiple such model variants with science-focused instructions and evaluated them on a scientific document reasoning benchmark for the usefulness of the retrieved document passages. Our findings suggest that models justify predictions in science tasks with fabricated evidence and leveraging scientific corpus as pretraining data does not alleviate the risk of evidence fabrication.


\end{abstract}

\section{Introduction}

Large Language Models (LLM) perform competitively in a majority of Natural Language Processing (NLP) tasks, but tend to hallucinate with seemingly plausible but misleading predictions~\cite{mallen2023not,jiang2023active}.
The black-box language models do not provide clear explanations or justifications for their predictions or decisions~\cite{mialon2023augmented}. In recent times, retrieval-augmented LMs help address these issues by augmenting the LLMs with non-parametric memory by employing neural retriever to extract relevant information from external knowledge resources such as document corpora~\cite{jiang2023active}.
Retrieving external knowledge helps the model to update with new knowledge, inject domain specific data, and memorize long-tail knowledge. 
Furthermore, these models are relatively small in the number of parameters and require less training and inference costs~\cite{borgeaud2022improving}.



While Retrieval-augmented LMs are shown to perform well on knowledge-intensive tasks~\cite{izacard2022few}, we have very limited understanding on their ability to perform on the science-focused downstream tasks.
For example, we can provide scientific documents as external knowledge at test time, and test the ability of the model to perform on science question and answering (QA) task.
In this setup, the model retrieves scientific documents relevant to the question, and then generates an answer conditioning on the retrieved documents.
Retrieved documents help the model predictions to be better interpreted and verified.
At the same time, we can assess the trustworthiness of these models to understand whether they justify the model predictions with accurate and relevant evidence~\cite{mallen2023not}.
Recognizing any failure modes is essential to ensuring the safe deployment, and avoiding potential risks or negative consequences of these models, specially across multiple scientific use cases and applications.

This work conducts a comprehensive evaluation on Retrieval-augmented LMs to improve our understanding of these models to perform on science tasks.
To this end, we used the ATLAS model architecture~\cite{izacard2022few} as an instance of the retrieval-based language model family to drive our experiments.
To adopt these models for science tasks, we provided a collection of scientific documents as external corpus during the model pretraining, instruction tuning and evaluation stages.
We evaluate the model performance on SciRepEval~\cite{singh2022scirepeval} benchmark to test whether model recognizes different scientific domains and disciplines from the given scientific documents.
Our hypothesis is that these models will be able to retrieve relevant information from scientific documents, integrate knowledge from diverse scientific domains, and reason over complex scientific concepts.
In particular, we evaluate the usefulness of retrieved passages in terms of their relevancy and diversity that support the model predictions.

\begin{figure*}[!t]
\centering
\begin{subfigure}{.48\textwidth}
  \centering
  \includegraphics[width=.98\linewidth]{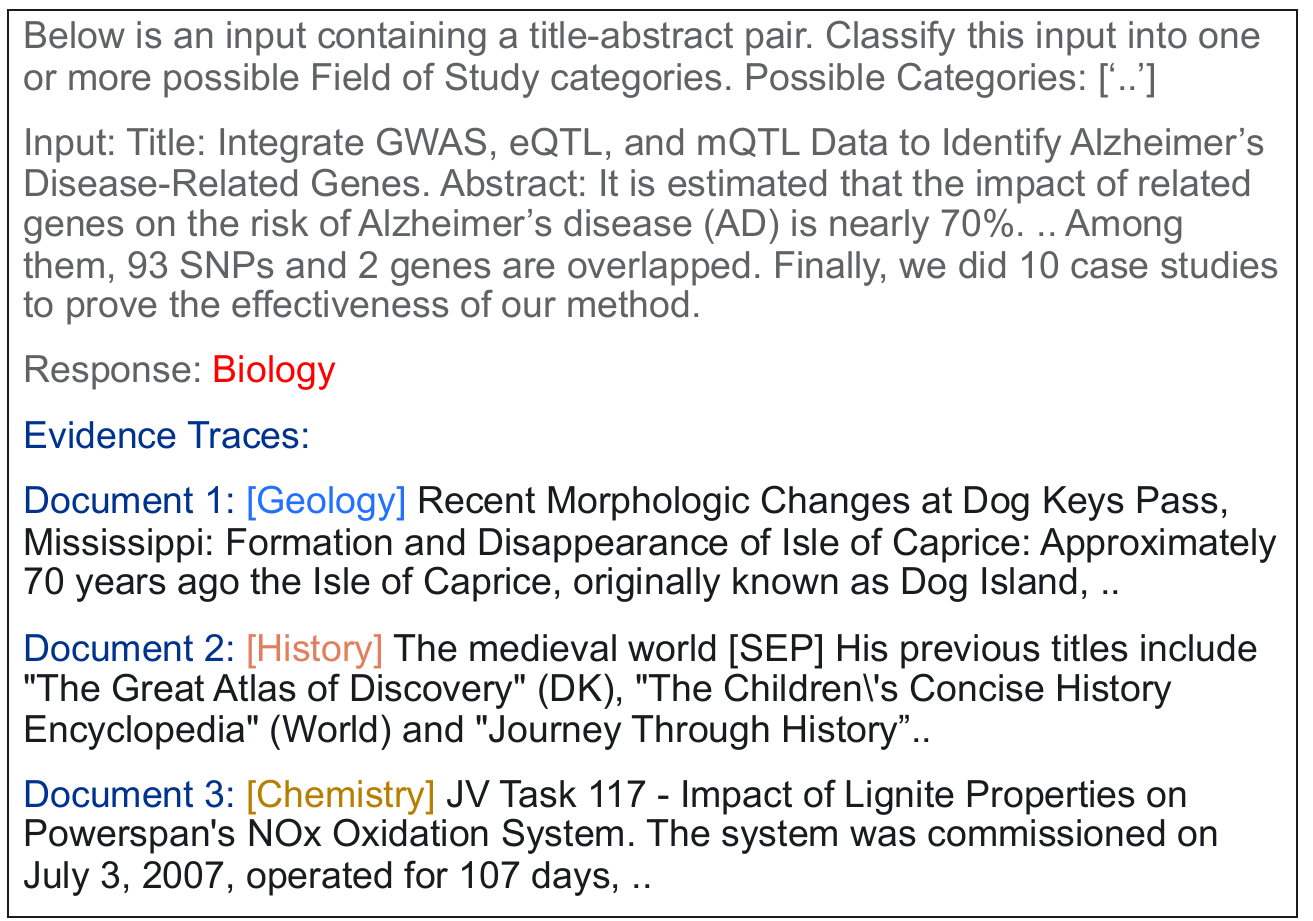}
\end{subfigure}
\begin{subfigure}{.48\textwidth}
  \centering
  \includegraphics[width=.98\linewidth]{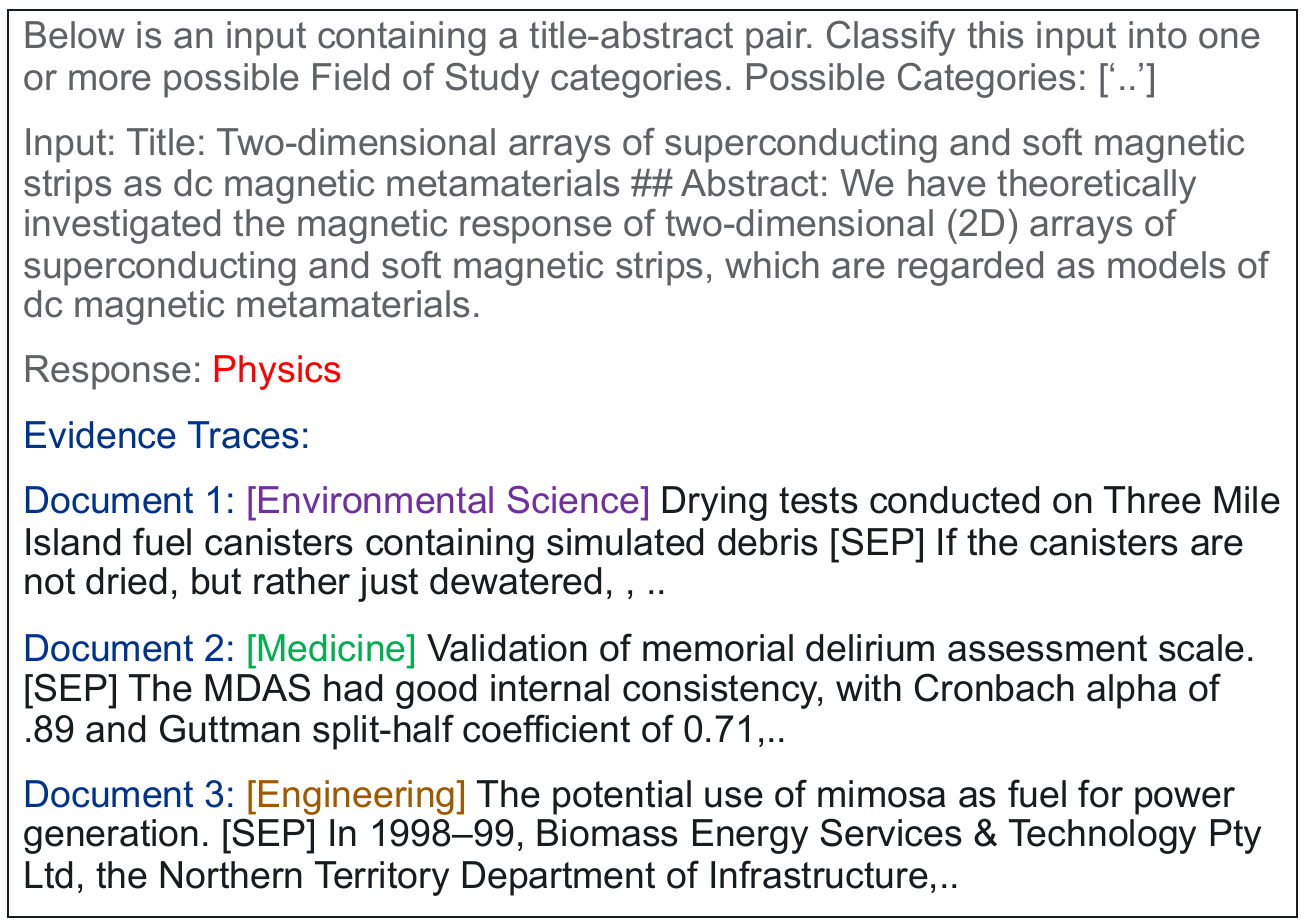}
\end{subfigure}
\caption{Example generations of the ATLAS instruction tuned model in the SciRepEval-FoS~\cite{singh2022scirepeval} task. We color the input query in gray, and the generated answer in red. We list three documents returned by the model as evidence to support the answer. We annotate each document by the corresponding scientific domain. For example, the model accurately generates the Biology domain that the passage belongs to, but justifies the answer with fabricated evidence as retrieved passages are in Geology, History and Chemistry.}
    \label{fig:atlas-example}
\end{figure*}

\section{Problem Formulation and Evaluation Setup}


\subsection{Problem Formulation}
Previous research on Retrieval Augmented LLMs focused on solving three major research questions: i) what to retrieve (e.g., chunks, tokens), ii) how to retrieve (e.g., input, intermediary and output layers), and iii) when to retrieve (e.g., once, every n$\geq$1 tokens).
A majority of proposed models such as REALM~\cite{guu2020retrieval}, DPR~\cite{karpukhin2020dense}, RAG~\cite{lewis2020retrieval}, and ATLAS~\cite{izacard2022few} retrieve text chunks and concatenate them in the input layer of the language model.
For example, ATLAS combines autoregressive text generation with retrieval-based language model pre-training based on the encoder-decoder architecture and fine-tuned on open-domain QA.

In this research, we aim to improve our understanding on the development of retrieval-based LMs for evidence extraction.
We focus on the following related research questions to drive our experiments.


\begin{description}
    \item[\namedlabel{RQ1}{(RQ1)}] How useful are the evidences generated from retrieval-augmented LLMs to justify model predictions in science tasks?
    \item[\namedlabel{RQ2}{(RQ2)}] How do the retrieval-augmented LLMs behave when provided with the scientific knowledge as the external document store? 
\end{description}

\subsection{Evaluation Setup}
\label{sec:experimental_setup}
In this section, we outline the datasets, models, benchmarks and metrics used in our experiments.

\paragraph{Scientific Text Datasets}
Retrieval Augmented LLMs provide ideal test bed for scientific applications since they can handle dynamic knowledge updates and different scientific domains and disciplines than what the models see during the pretraining.
We focus on evaluating the Retrieval Augmented LLMs on their ability to understand scientific language and retrieve from multiple scientific knowledge sources.
We preprocess the S2ORC~\cite{lo2019s2orc} dataset to create a collection of 354M text passages.
Each passage has a maximum of 512 tokens, or 100 words that are concatenated with the corresponding title of the document the passage belongs to.
We record 19 different scientific domains in the S2ORC collection\footnote{S2ORC dataset covers 19 scientific domains; 
Art,
Philosophy,
Political-Science,
Sociology,
Psychology,
Geography,
History,
Business,
Economics,
Geology,
Physics,
Chemistry,
Biology,
Mathematics,
Computer Science,
Engineering,
Environmental science,
Material science,
Medicine
}.

\begin{figure*}[!t]
    \centering
    \includegraphics[scale=0.35]{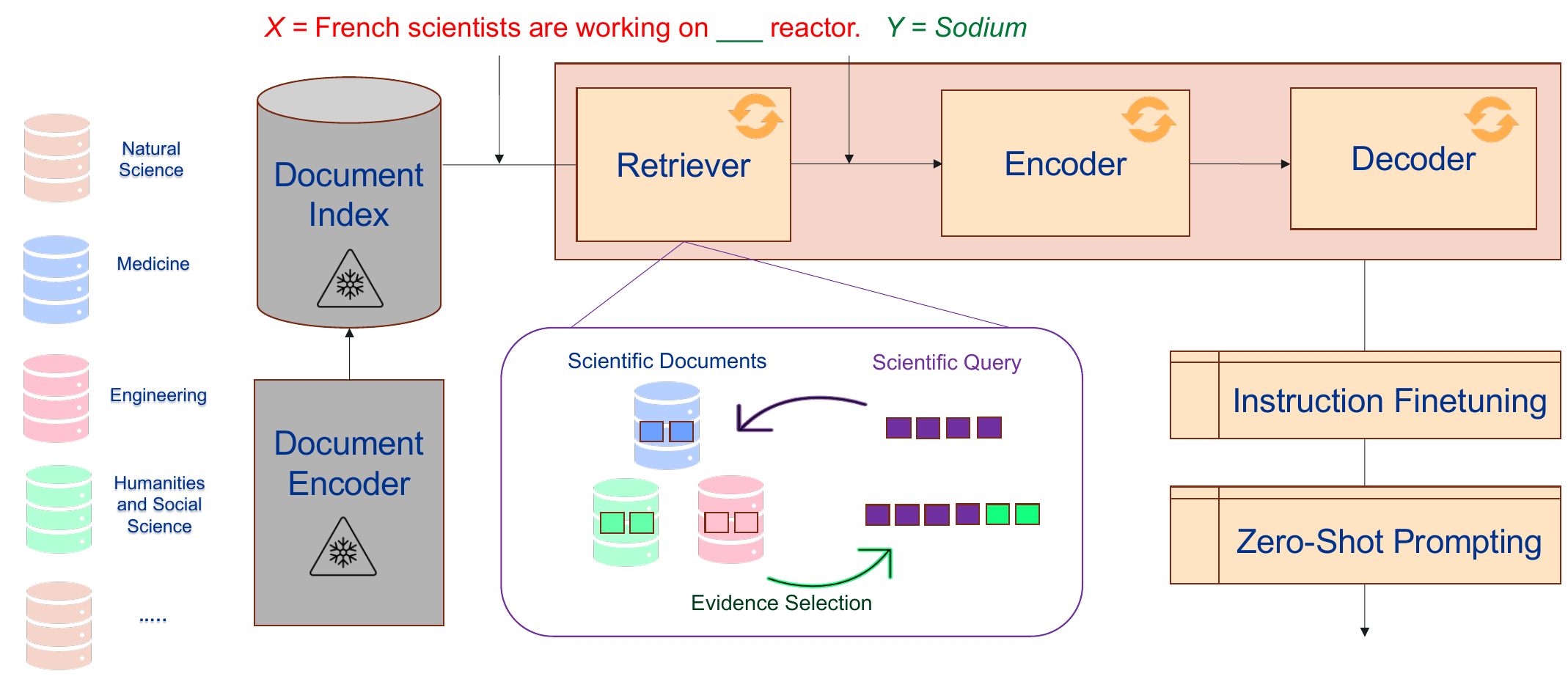}
    \caption{Experimental setup to measure the effectiveness of retrieval augmented LLMs in scientific document reasoning tasks. We trained the ATLAS-Science model to evaluate for scientific document reasoning tasks. We use the FoS task data in the SciRepEval benchmark to perform instruction tuning and evaluate the model in both in-distribution (FoS) and out-of-distribution (MAG) tasks with zero demonstrations during the inference.}
    \label{fig:sciguide}
\end{figure*}

\paragraph{Models} Our experiments are based on ATLAS (220M)~\cite{izacard2022few} model architecture unless explicitly mentioned.
ATLAS uses the Fusion-in-decoder architecture to fuse the retrieved text chunks with the input queries during the pretraining.
In addition to the ATLAS model pretrained with common crawl (CC) and Wikipedia, we also train ATLAS-Science (220M) model from scratch with the S2ORC scientific text datasets.
For a fair comparison with ATLAS, we initialize the ATLAS-Science model with the \textit{T5-lm-adapt}~\cite{raffel2020exploring} model and trained jointly with retrieval model, \textit{Contriever}~\cite{izacard2021unsupervised}.
Figure~\ref{fig:sciguide} shows the overview of different components used in the ATLAS-Science model.
We provide the collection of scientific text passages as external retrieval corpus.
First, we encode the scientific text passages (354M) with the \textit{Contriever} model, and construct a document index in the FLAT~\cite{izacard2022few} mode for faster retrieval. 
Second, we use the same passages for model pretraining and ensure that the passages used for pretraining are distinct from passages used to build the document index.
Third, we train the retriever with the \textit{query side finetuning} approach that originally introduced in the ATLAS model.
This approach is very efficient in model training since it keeps the document encoder frozen while training the parameters corresponding to the query encoder (Figure~\ref{fig:sciguide}).
All the models are trained for the same number of tokens for a fair comparison.

\begin{table*}[!t]
\centering
\caption{Summary of different pretraining, instruction tuning and benchmark datasets used across T5 and ATLAS models. We report the performance of the standalone LLM i) T5 (pretrained with C4), retrieval-augmented LLMs, ii) ATLAS model (pretrained with CC and Wikipedia) and iii) ATLAS-Science  model (pretrained with S2ORC) text datasets. We used the S2ORC dataset as the external retrieval corpus in the instruction tuning and evaluation stages to make a fair comparison.}
\label{tab:model_setup}
\scalebox{0.85}{
\begin{tabular}{|c|c|c|c|c|c|c|}
\hline
\multirow{2}{*}{\textbf{Model}} & \multicolumn{2}{c|}{\textbf{Pretraining}}                                   & \multicolumn{2}{c|}{\textbf{Instruction Tuning}}                      & \multicolumn{2}{c|}{\textbf{Evaluation}}                       \\ \cline{2-7} 
                                & \textbf{Data}              & \textbf{Retrieval Corpus} & \textbf{Data}        & \textbf{Retrieval Corpus} & \textbf{Data} & \textbf{Retrieval Corpus} \\ \hline \hline
\multirow{2}{*}{T5}          & \multirow{2}{*}{C4} & \multirow{2}{*}{N/A}     & \multirow{6}{*}{FOS} & \multirow{2}{*}{N/A}    & FOS           & \multirow{2}{*}{N/A}    \\ \cline{6-6}
&                            &                           &                      &                           & MAG           &                           \\ \cline{1-3} \cline{5-7}
\multirow{2}{*}{ATLAS}          & \multirow{2}{*}{CC + Wiki} & \multirow{2}{*}{Wiki}     &  & \multirow{4}{*}{S2ORC}    & FOS           & \multirow{4}{*}{S2ORC}    \\ \cline{6-6}
                                &                            &                           &                      &                           & MAG           &                           \\ \cline{1-3}\cline{6-6}
\multirow{2}{*}{ATLAS-Science}      & \multirow{2}{*}{S2ORC}    & \multirow{2}{*}{S2ORC}   &  &    & FOS           &    \\ \cline{6-6}
                                &                            &                           &                      &                           & MAG          &                           \\ \hline
\end{tabular}
}
\end{table*}

\paragraph{Instruction Tuning}
We use the SciRepEval~\cite{singh2022scirepeval} benchmark for training and evaluating the models for the scientific evidence extraction.
SciRepEval provides 25 challenging tasks across four formats: classification, regression, ranking, and search.
In this work, we focus on the classification formatted tasks, \textit{Fields of study (FoS)} and \textit{MAG} due to two main reasons.
First, we need benchmark tasks that test the ability of the models to understand diverse scientific domains and disciplines\footnote{FoS tasks include instructions from following domains;  Materials science,  Economics,  Chemistry,  Medicine,  Psychology,  Geography,  Geology,  Political science,  Engineering,  Philosophy,  Sociology,  Physics,  Computer science,  Law,  History,  Biology,  Agricultural and Food sciences,  Environmental science,  Business,  Education,  Art,  Linguistics,  Mathematics}.
For example, FoS task tests the ability of the model to recognize which domain the given text passage belongs to.
Second, we want to evaluate on specific instruction template to avoid any prompting bias.

Previous research~\cite{izacard2022few} has shown that ATLAS model is able to learn knowledge intensive tasks with very few training examples (aka few shot learning).
To allow the model to perform on the downstream tasks, we tune the model with scientific instructions.
We design an instruction template\footnote{Instruction template for FoS and MAG tasks is "\#\#\# Below is an input containing a title-abstract pair. Classify this input into one or more possible Field of Study categories. \#\#\# Possible Categories: [..] \#\#\# Input: \#\# Title: .. \#\# Response:"} to guide the model to generate the scientific domain that each passage belongs to.
We tune the model with \textit{Fields of study (FoS)} training data after converting them to instructions.
This process resulted 541,218 training instructions that used to perform instruction tuning.
For a fair comparison, we tune the T5 and ATLAS models with these instructions.
In comparison to the T5 model, ATLAS models retrieve the top-k relevant passages from the S2ORC document store to augment the instruction tuning process.
There are 68,147 and 3,751 test instructions in the FoS and MAG tasks, respectively.
We use MAG instructions to test the out-of-distribution task performance.
Table~\ref{tab:model_setup} summarizes the pretraining, instruction tuning and evaluation data used for the ATLAS and ATLAS-Science models.

\paragraph{Metrics}
We use the Exact Match (EM) and F1-Score to evaluate the task accuracy.
EM metric evaluates the exact token overlap between the ground truth and generated answers.
In addition, we design two metrics to evaluate the relevance and diversity of the extracted evidence: the \textit{relevance} and \textit{diversity} metrics. The relevance metric calculates the ratio of the domains present in Top-k evidences matching with the scientific domain corresponding to the query.
The diversity metric calculates the ratio of the unique evidences in comparison to the total evidences.
Both metrics are in the range of zero and one, with higher the metric scores, higher the quality in the generated evidences.

\section{Measuring the Effectiveness of the Scientific Evidence Extraction}
In this section, we address~\ref{RQ1} and~\ref{RQ2} by evaluating the usefulness of the evidence extracted from retrieval-augmented LLMs in performing science tasks.
We test how retrieval-augmented LLMs behave when provided the scientific knowledge as external memory.

\paragraph{Retrieval-augmented LLMs justify model predictions in science tasks with fabricated evidence}
We evaluate the pretrained ATLAS model~\cite{izacard2022few} on the benchmark tasks \textit{Fields of study (FoS)} and \textit{MAG}. 
Additionally, we tune ATLAS model with the \textit{Fields of study (FoS)} instructions (as described in Section~\ref{sec:experimental_setup}).
We use the S2ORC document index to evaluate the instructions tuned ATLAS model in the zero-shot prompting stage. 
We report the performance of ATLAS model in Table \ref{Tab:1}. 
First, we observe that the accuracy of the ATLAS model is better than that of T5 in both the tasks, demonstrating the importance of retrieval augmentation. 
Second, we observe that although the ATLAS model has achieved a acceptable accuracy of $84.42$\% in \textit{FOS} and $59.10$\% in \textit{MAG}, the retrieved evidences are extremely poor in terms of relevance to the query topic.
The model only achieves $0.06$ relevance score suggesting that the passages returned by the model as evidences do not align with the domain of the query.
For example, ATLAS model returns the passages in Geology, History and Chemistry as evidence for a Biology query as shown in Figure~\ref{fig:atlas-example}.
The retrieved passages are not at all related to the corresponding query topics, rendering them useless.
Finally, we also evaluate the faithfulness of the retrieved docs in terms of diversity score.
This score is very low, suggesting that the evidences remain similar across all test queries.
Our qualitative check suggests that the top-20 passages returned by the model for different queries are exactly same, while the generated answers are different and mostly accurate in comparison to the ground truth.
These observations suggest that the ATLAS model fabricates the evidence to justify the model predictions. 

\paragraph{Scientific knowledge provided as pretraining data does not alleviate the evidence fabrication}
To explore the impact of the pretraining data on downstream scientific tasks, we repeat the evaluation with the \textit{ATLAS-Science} model (as described in Section~\ref{sec:experimental_setup}). 
Note that the ATLAS-Science model is pretrained from scratch with S2ORC scientific text data provided as both pretraining and external document store.
We evaluate the ATLAS-Science on two benchmark tasks. 
The results are tabulated in third row of Table \ref{Tab:1}. 
In comparison to ATLAS, the accuracy of ATLAS-Science model has a slight improvement in \textit{FOS}, whereas it depreciates in \textit{MAG}. 
Thus we see that scientific corpus does not help much in improving the performance of the model. 
More importantly, the relevance and diversity of the retrieved passages only slightly improves over ATLAS. 
This indicates that leveraging scientific corpus as pretraining data is not an effective approach to address the challenge of evidence fabrication. 

\begin{table*}[!t]
\centering
\caption{Model ablation study to evaluate performance on in-distribution (SciDocs-FoS) and out-of-distribution (SciDocs-MAG) field of study instruction tuning datasets.}
\label{Tab:1}
\scalebox{0.90}{
\begin{tabular}{|c|r|r|r|r|r|r|r|r|}
\hline
\multirow{3}{*}{\textbf{Model}}  & \multicolumn{4}{c|}{\textbf{In-distribution Performance}}                                                                                                                                                                                                                                    & \multicolumn{4}{c|}{\textbf{Out-of-distribution Performance}}                                                                                                                                                                                                                                    \\ \cline{2-9} 
                                &  \multicolumn{2}{c|}{\textbf{Accuracy}} & \multicolumn{2}{c|}{\textbf{Evidence Generation}}                                                                            & \multicolumn{2}{c|}{\textbf{Accuracy}} & \multicolumn{2}{l|}{\textbf{Evidence Generation}}                                                                            \\ \cline{2-9} 
                                & \textbf{EM}     & \textbf{F1}   & \textbf{Relevance} & \textbf{Diversity} & \textbf{EM}     & \textbf{F1}    & \textbf{Relevance} & \textbf{Diversity} \\ \hline
\textbf{T5}                & 83.33                   & 0.87              &                                                                N/A      & N/A                                                                     &   57.90                      &   0.72                  &   N/A                                                                       &    N/A                                                                     \\ \hline
\textbf{ATLAS}        & 84.42                   & 0.92             & 0.06                                                                  &      5E-5                                                                  & 59.10                       & 0.75                  & 0.07                                                                        & 60E-5                                                 \\ \hline
\textbf{ATLAS-Science}    & 84.70                  & 0.92            & 0.05                                                                  &        8E-5                                                                 & 57.80                     & 0.73                    & 0.05                                                                        &      100E-5                                          \\ \hline
\end{tabular}
}
\end{table*}
\section{Related Work}
LLMs can be augmented with various external modules such as document corpus \cite{gao2020pile}, vector databases\footnote{https://github.com/jerryjliu/llama\_index}, etc. 
Typically, the augmentation is accomplished in two ways, namely sparse (such as Bag of words) and dense where Neural network is used to encode documents. 
Dense retrievers are widely used in present time mainly due to the good representation capability of neural networks. 
Recent works suggest that the retrieval-augmented LLMs has significant improvement over the standard LLMs across various tasks especially with respect to scale~\cite{guu2020retrieval,lewis2020retrieval}. 
REALM and RAG  are the initial efforts where they train the retriever and language model by representing documents as
latent variable, and minimizing the language model objective \cite{guu2020retrieval,lewis2020retrieval}. 
REALM leverages masked-language modeling as an objective to pretrain the model in end to end fashion. 
However, it is computationally very expensive to train these models that requires to retrain the entire index with new knowledge. 
Guu et al.~\cite{guu2020retrieval} explored the concept of query-side finetuning that only refreshes the query encoder whereas document encoder remains frozen. Izacard et al.~\cite{izacard2020leveraging, izacard2021unsupervised} proposed various ways to improve the retrieval augmented models, including novel learning objectives to align retriever with the language model~\cite{izacard2021unsupervised}. 
Furthermore, RETRO~\cite{ borgeaud2022improving} shows the benefits of scaling the retrieval memory to trillions of tokens. 
ATLAS~\cite{izacard2022few} experiment with various design and training configurations for retrieval augmented models with a focus on few shot learning ability. 
\section{Conclusion}
In this study, we explored the efficacy of retrieval augmented language models on science tasks. 
Our experiments were based on ATLAS model which is a state of the art retrieval augmented language model with few shot capability. 
We performed a systematic evaluation on the performance of different ATLAS model variants in two scientific document reasoning tasks.  
Our experiments on the pretrained ATLAS model reveal that the model demonstrates acceptable performance in science tasks but the evidences are fabricated. 
We also observe that pretraining the model with scientific corpus does not alleviate evidence fabrication. We plan to develop techniques to alleviate these issues in a future work.
\section*{Acknowledgements}
This work was supported by the NNSA Office of Defense Nuclear Nonproliferation Research and Development, U.S. Department of Energy, and Pacific Northwest National Laboratory, which is operated by Battelle Memorial Institute for the U.S. Department of Energy under Contract DE-AC05–76RLO1830. This article has been cleared by PNNL for public release as PNNL-SA-189029.

\bibliography{main.bib}
\bibliographystyle{abbrv}

\end{document}